\documentclass{article}

     \usepackage[preprint]{neurips_2021}

\usepackage[utf8]{inputenc} 
\usepackage[T1]{fontenc}    
\usepackage{hyperref}       
\usepackage{url}            
\usepackage{booktabs}       
\usepackage{amsfonts}       
\usepackage{nicefrac}       
\usepackage{microtype}      
\usepackage{xcolor}         

\usepackage{amsmath}
\usepackage{times}
\usepackage{graphicx}
\usepackage{color}
\usepackage{multirow}

\graphicspath{ {figures/} }

\title{Spontaneous Emerging Preference in Two-tower Language Model}

\author{%
  Zhengqi He  \\
  Lab for Neural Computation and Adaptation, RIKEN Center for Brain Science, Japan\\
  \texttt{zhengqi.he@riken.jp} \\
   \And
   Taro Toyoizumi \\
  Lab for Neural Computation and Adaptation, RIKEN Center for Brain Science, Japan \\
Department of Mathematical Informatics, Graduate School of Information Science \\
and Technology, The University of Tokyo\\
  \texttt{taro.toyoizumi@riken.jp} \\
}

\begin{document}

\maketitle

\begin{abstract}
  
The ever-growing size of the foundation language model has brought significant performance gains in various types of downstream tasks. With the existence of side-effects brought about by the large size of the foundation language model such as deployment cost, availability issues, and environmental cost, there is some interest in exploring other possible directions, such as a divide-and-conquer scheme. In this paper, we are asking a basic question: are language processes naturally dividable? We study this problem with a simple two-tower language model setting, where two language models with identical configurations are trained side-by-side cooperatively. With this setting, we discover the spontaneous emerging preference phenomenon, where some of the tokens are consistently better predicted by one tower while others by another tower. This phenomenon is qualitatively stable, regardless of model configuration and type, suggesting this as an intrinsic property of natural language. This study suggests that interesting properties of natural language are still waiting to be discovered, which may aid the future development of natural language processing techniques.
  
\end{abstract}

\section{Introduction}

Recently, AI is witnessing a rising trend of foundation models \citep{bommasani2021opportunities}, which are generally defined as models trained on broad data, usually unsupervised. The trained model can then be used to solve other downstream tasks. In the field of natural language processing, the famous foundation model examples include ELMo \citep{peters-etal-2018-deep}, GPT \citep{radford2018improving} and Bert \citep{devlin2018bert}. As has been noticed, continuously increasing the scale of model size and data for training boosts model performance. This phenomenon is described by a scaling law \citep{kaplan2020scaling}. As a result, we see a trend of language models with larger sizes and better performance, such as T5 \citep{raffel2020exploring}, GPT3 \citep{brown2020language} Megatron-Turing NLG \citep{smith2022using}. Although a larger language model brings significant performance gains in downstream tasks, their enormous sizes also bring problems, such as deployment cost \citep{sanh2019distilbert}, availability issue \citep{izsak2021train}, and environmental cost \citep{schwartz2020green, patterson2021carbon}. 

There are lots of attempts aimed at solving these issues and making language models lighter and smaller. Available techniques include knowledge distilling \citep{sanh2019distilbert}, parameter sharing \citep{lan2019albert}, weight pruning \citep{michel2019sixteen}, reduced precision training \citep{micikevicius2017mixed}, quantization \citep{zafrir2019q8bert} and adaptive methods \citep{hou2020dynabert}, to name a few. One relative under-explored strategy is the divide-and-conquer way of building a language model. One effective method in this direction is mixture-of-experts language modeling. Famous works include Switch Transformers \citep{fedus2021switch} and Glam \citep{du2022glam}. Both rely on a mixture of experts who are usually sparsely activated. Compared to dense models, this design is capable of scaling up the parameters with reduced computational and memory resources. Some methods also explicitly try to split syntax from semantics \citep{russin2019compositional, havrylov2019cooperative} and demonstrate performance gain on some downstream tasks.

One key question we can ask about the divide-and-conquer method of language modeling is, are language processes naturally dividable? As a first step towards studying this problem, we look at a simple scenario where two identical language models are trained side-by-side. We call this setting the "two-tower language model." A preliminary study of this setting suggests a positive answer to our question. We discover a stable pattern of self-emergence of modeling preference in the two-tower language model. In other words, one subgroup of language features is better modeled by one tower, whereas the other subgroup is better modeled by the other tower. The emerging pattern of preference is qualitatively stable across different language model types. Thus, this phenomenon appears to be an inherent property of language rather than a result of the modeling process. Therefore, our main contribution is the discovery of the self-emergence preference effect in a two-tower language model, indicating the existence of dividable structures to process natural language. 

\section{Related Works}

One line of related work is language modeling with deep neural networks (DNN). DNN-based language models are usually trained to do self-prediction on a large corpus of unlabeled language data. Common strategies to do self-prediction include cloze style, where each word is predicted when given a bi-lateral context; auto-regressive style, where each word is predicted when given the previous context; and masked language modeling (MLM) style, where a certain percentage of words are masked and predicted by remaining un-masked words. Examples of language models using cloze, auto-regressive and MLM style are ELMo \citep{peters-etal-2018-deep}, GPT \citep{radford2018improving} and Bert \citep{devlin2018bert} respectively. Language models using a hybrid approach are also available, such as XLNet \citep{yang2019xlnet}, a model that combines auto-regressive and masked language modeling. 

Other related works include language models with a two-tower modeling strategy. We can roughly divide the models into two groups, symmetrical two-tower models and asymmetrical two-tower models. Symmetrical two-tower models have two symmetrical sub-units which work together to perform certain tasks. One example is the two-tower GPT \citep{baevski2019cloze} language model, which performs cloze-style language modeling. It has one GPT handling forward context and another GPT handling backward context. Another example is mirror-Bert \citep{liu2021fast}, where two parameter-shared Berts are fine-tuned together to refine the learned sentence representation via contrastive learning. The asymmetrical two-tower model consists of two sub-units with different configurations. For example, Electra \citep{clark2020electra} has two MLM language models. A smaller generator language model generates word replacement and a bigger discriminator language model predicts if the word has been replaced. /cite[russin2019compositional] is another example, in which language information is explicitly divided into syntax and semantics streams, with syntax information serving as an attention over semantics information.

\section{Method}

\subsection{Two-tower Language Model}

\begin{figure}[t]
\centering
\includegraphics[width=0.6\linewidth]{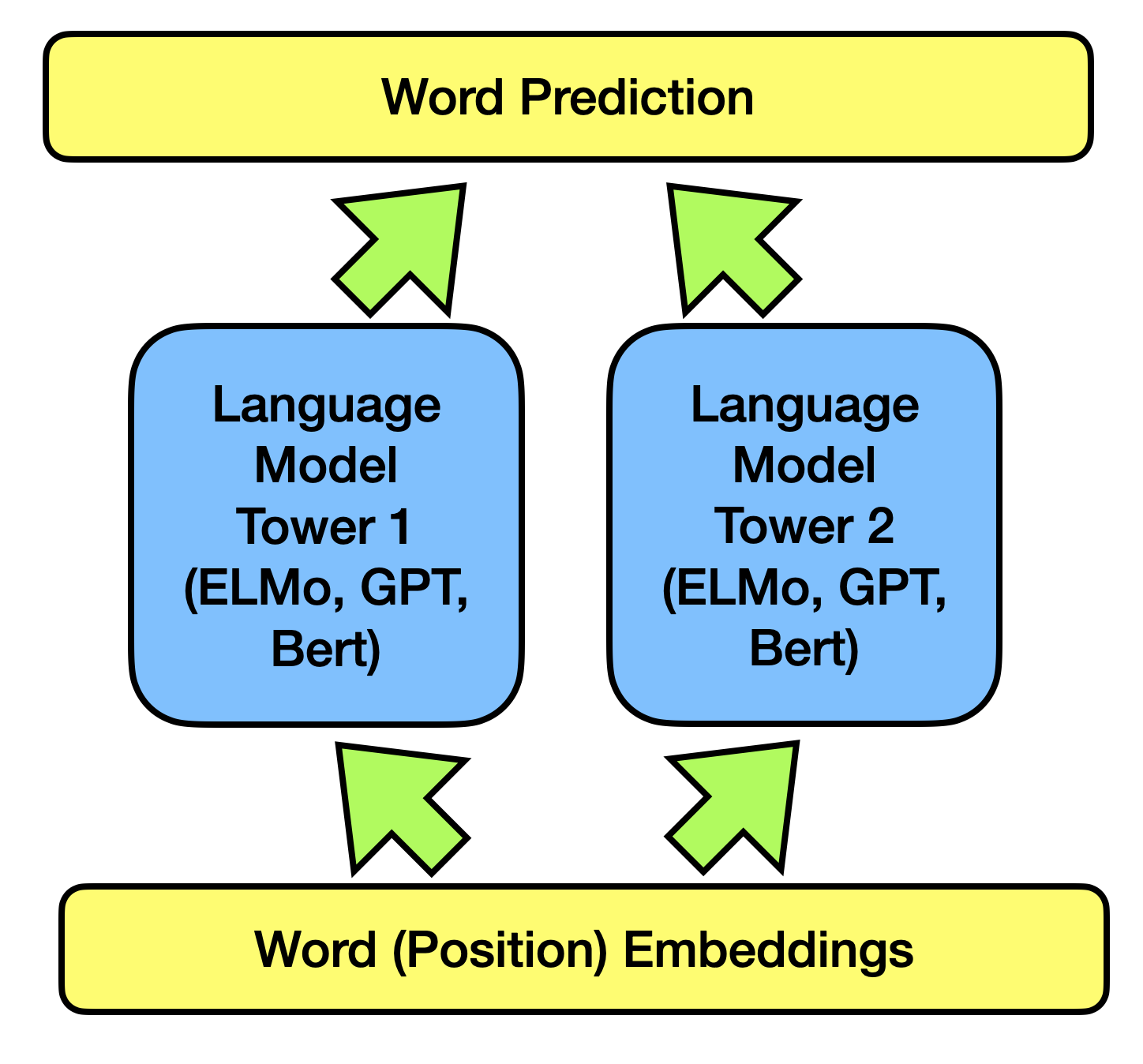}
\caption{Network topology of the two-tower language model. Word vectors of a sequence of tokens are encoded separately by two language models. The language model can be either ELMo, GPT, or Bert style. The encoded representations are then concatenated and do cloze/auto-regressive/MLM style self-prediction tasks.}
\label{2tower}
\end{figure}

The network topology of our two-tower language model is shown in Fig. \ref{2tower}. Assume a sequence of tokens $ T = [t_1, t_2, ... , t_n]$ going through a word embedding layer and form a sequence of word vectors $ X = [x_1, x_2, ... , x_n]$. (Note that Positional embedding would be added for Bert and GPT cases.) Then, the word vectors would go through two encoders with the same configuration $f_1, f_2$ separately, and form two sequences of hidden representations $h_{1i} = f_{1i}(x)$, $h_{2i} = f_2(x)_i$, where $i$ represents word position index. We choose three types of neural networks for language encoders $f_1, f_2$. For case 1 we choose the ELMo style, where multiple layers of bidirectional LSTMs \citep{hochreiter1997long, graves2013speech} are utilized as the neural network backbone. For case 2 we choose the GPT style and for case 3 we choose the Bert style. For both the GPT style and the Bert style, transformers \citep{vaswani2017attention} are selected as the neural network backbone. 

Depending on the network type, token visibility at each position differs for each hidden representation. For ELMo, where cloze style self-prediction is used, the $i$-th hidden representation would be able to access both the previous and following context. The hidden representation can be represented as $h_{(1,2)i} = \mathrm{ELMo}_{(1,2)}(x_1,...,x_{i-1},x_{i+1}, ..., x_n)$. For GPT, where auto-regressive style self-prediction is employed, the $i$-th hidden representation would only be able to see the previous context. The hidden representation can then be written as $h_{(1,2)i} = \mathrm{GPT}_{(1,2)}(x_1,...,x_{i-1})$. For Bert, where MLM style self-prediction is used, the $i$-th hidden representation would be able to see the whole context except for masked tokens. The hidden representation can then be written as $h_{(1,2)i} = \mathrm{Bert}_{(1,2)}(x_1,...,\tilde{x}_i,...,x_n)$, where $i$ represents the index for masked tokens only. $\tilde{x}_i$ represents the word vector of a special token [MASK] at position $i$.

Then, hidden representations coming from both towers are concatenated $h_i = \mathrm{cat} (h_{1i}, h_{2i})$. After concatenating the representations, the probability at each word position can be calculated as follows:

\begin{equation} \label{prob}
\begin{split}
&p_i(t|t_{(visible, i)}) = \mathrm{exp}(x(t)^T h_i) / \sum_{a} \mathrm{exp}(x_a^T h_i)
\end{split}
\end{equation} 

where $t_{(visible, i)}$ means visible tokens for position $i$, namely $t_1,...,t_{i-1},t_{i+1}, ..., t_n$ for ELMo case, $t_1,...,t_{i-1}$ for GPT case and $t_1,...,t_n$ for Bert case. And for Bert, $i$ is for masked tokens only. $x_a$ represents embedding vectors of all the words in our vocabulary, and $a$ represents the token index in our vocabulary.

\subsection{Recognizing Preference Pattern}

\begin{figure}[t]
\centering
\includegraphics[width=0.6\linewidth]{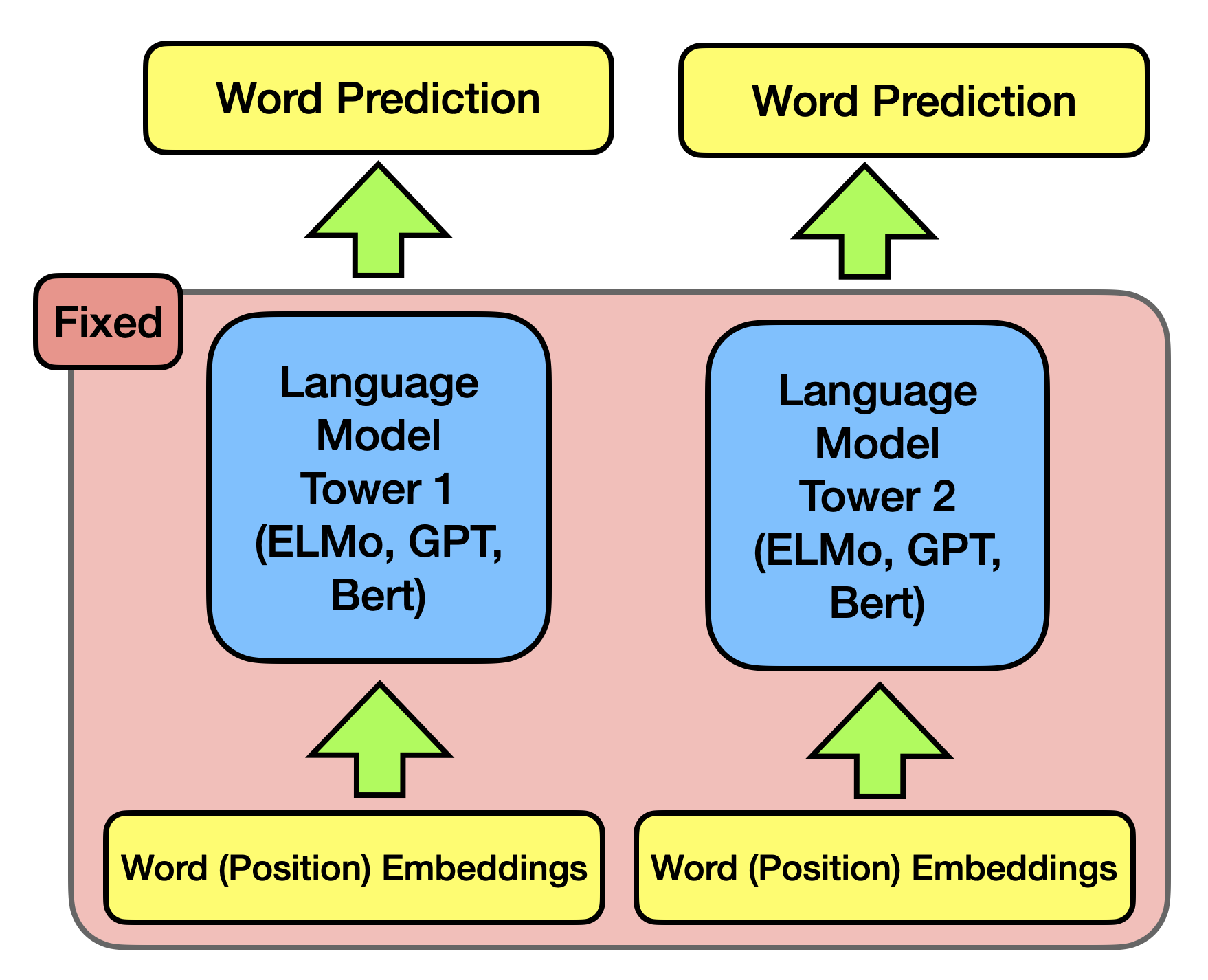}
\caption{Network configuration to study functional preference of the two-tower language model after the previous language modeling step. The parameters of word embedding layers and language models are fixed, and only output layers are retrained to do language modeling. }
\label{2tower_s2}
\end{figure}

As the two-tower language models start with symmetrical configuration and random initialization, they should start with no preference over any specific language function. In order to test whether any preference pattern emerges following the training of the two-tower language model, we fix the parameters of the embedding layers and the language model part and retrain the output prediction layers separately for each tower.  The network configuration for this step is shown in Fig. \ref{2tower_s2}. 

Then we'll have the probability of each word calculated separately from each tower:

\begin{equation} \label{prob12}
\begin{split}
& p_{i1}(t|t_{(visible, i)}) = \mathrm{exp}(x(t)^T h_{1i}) / \sum_{a} \mathrm{exp}(x_a^T h_{1i}) \\
& p_{i2}(t|t_{(visible, i)}) = \mathrm{exp}(x(t)^T h_{2i}) / \sum_{a} \mathrm{exp}(x_a^T h_{2i})
\end{split}
\end{equation} 

One thing we don't yet define is which tower should be tower 1 and which tower should be tower 2. Here, we define tower 1 and tower 2 so that:

\begin{equation} \label{t1t2}
\displaystyle \mathop{\mathbb{E}}_{t \in C} p_{i1}(t|t_{(visible, i)}) \geq \displaystyle \mathop{\mathbb{E}}_{t \in C} p_{i2}(t|t_{(visible, i)})
\end{equation} 

$C$ is the collection of all the words in our corpus. Thus, the order of the tower is defined so that the overall average word prediction probability for tower 1 is always greater than or equal to tower 2. Here, we name tower 1 "the primary tower", and tower 2 "the secondary tower". Hence, we can define the measure of the token preference score as the difference of average log probability between two separately trained model towers, which is described as:

\begin{equation} \label{pref_score}
\begin{split}
& s_a = \log \Bar{p}_{1a} - \log \Bar{p}_{2a} \\
& \Bar{p}_{(1,2)a} = \displaystyle \mathop{\mathbb{E}}_{t\in C} (1/n)\sum_{i=1}^n p_{i(1,2)}(t=t_{a}|t_{(visible, i)})
\end{split}
\end{equation} 

where $s_a$ is the preference score of the primary tower over the secondary tower for word $a$ and $t_a$ represents the a-th token in our vocabulary.

\section{Experiment Setup}

\subsection{Data}

Since our goal is to study self-emerging preferences in two-tower language models, we perform only the language modeling step without considering downstream tasks. Thus, only unlabeled language corpus data is needed. We use the same dataset as Bert, which is the BookCorpus \citep{zhu2015aligning} and English Wikipedia, around 3 billion words in total. We use word piece tokenization, also the same as Bert. The total number of unique tokens in our vocabulary is around 30K. The same dataset, tokenizer, and vocabulary are employed for ELMo/GPT/Bert cases. 

\subsection{Model}

We study three types of language models, namely ELMo, GPT, and Bert. For Bert model, we study two different network configurations, namely Bert-base and Bert-large. The network configuration of each type that we've experimented with is shown in the Table \ref{network} below.

\begin{table*}[h]
\caption{Network configurations.}
\label{network}
\begin{center}
\begin{tabular}{|l|cccc|}\hline
Model Type       & ELMo &	GPT &	Bert-base &	Bert-large\\
\hline
Number of layers & 2    &   12  &  12         &        24 \\
Hidden size      & 4096 &   768 &  384        &       512 \\
Intermediate size&  -   &   3072&  1024       &       2048\\
Attention heads  &  -   &   12  &  12         &        16 \\

\hline
\end{tabular}
\end{center}
\end{table*}

For ELMo, we combine two LSTMs, one for the forward direction and one for the backward direction. We directly use the Pytorch \citep{paszke2019pytorch} implementation of the LSTM. The input token is shifted left or right by one position for the forward and backward LSTM. This is to make sure that the output encoding of a specific token would not see itself. For GPT, we use Hugging Face \citep{wolf2019huggingface} implementation of GPT2. Since the token has been shifted inside the model, we won't need to handle it explicitly. For Bert, we use Deepspeed \citep{rasley2020deepspeed} implementation of Bing Bert. We make use of the highly optimized Deepspeed transformer kernel for better performance. 

\subsection{Training Details}
For training software, we use the Deepspeed software package. Deepspeed offers some features useful to our project. For example, Deepspeed provides an easy interface for distributed training, so that we can easily spread out the training load onto 6 Nvidia Tesla V100 GPUs. It also provides a micro-batch management option to simulate large batch-size training to reduce computation and communication costs. It also supports a half-precision option with a fused Adam \citep{kingma2014adam} optimizer compatible with the half-precision optimization. For the Adam optimizer parameter, we choose $\beta_1 = 0.9$, $\beta_2 = 0.98$, weight decay of 0.01, learning rate 1e-3. Instead of training the models until they converge, we follow the idea proposed in 24hBert \citep{izsak2021train} and control the training time of each case within approximately one day.

\section{Result}

\subsection{Correlation of Token Preference Score}

After training a two-tower language model, we can calculate the preference score $s_j$ for each token using Eq. \ref{pref_score}. Since the two towers share network configuration with random parameter initialization, we expect $s_j$ to be somehow random. However, after training several instances of two-tower language models, we notice that the preference score $s_j$ follows some self-emerged pattern which is qualitatively the same for all the three language model types we've tested.

We calculate $s_j$ for each token and each instance of the two-tower language model and form a vector $s = [s_1, s_2, ..., s_N]$, where $N$ is the size of our vocabulary. The elements are usually sorted in descending order according to token frequency. By comparing vector s among different trained two-tower language model instances, we can have an idea of how consistent $s$ can be. The consistency can be simply measured by the correlation between two s vectors of two model instances.

\begin{figure}[t]
\centering
\includegraphics[width=1.0\linewidth]{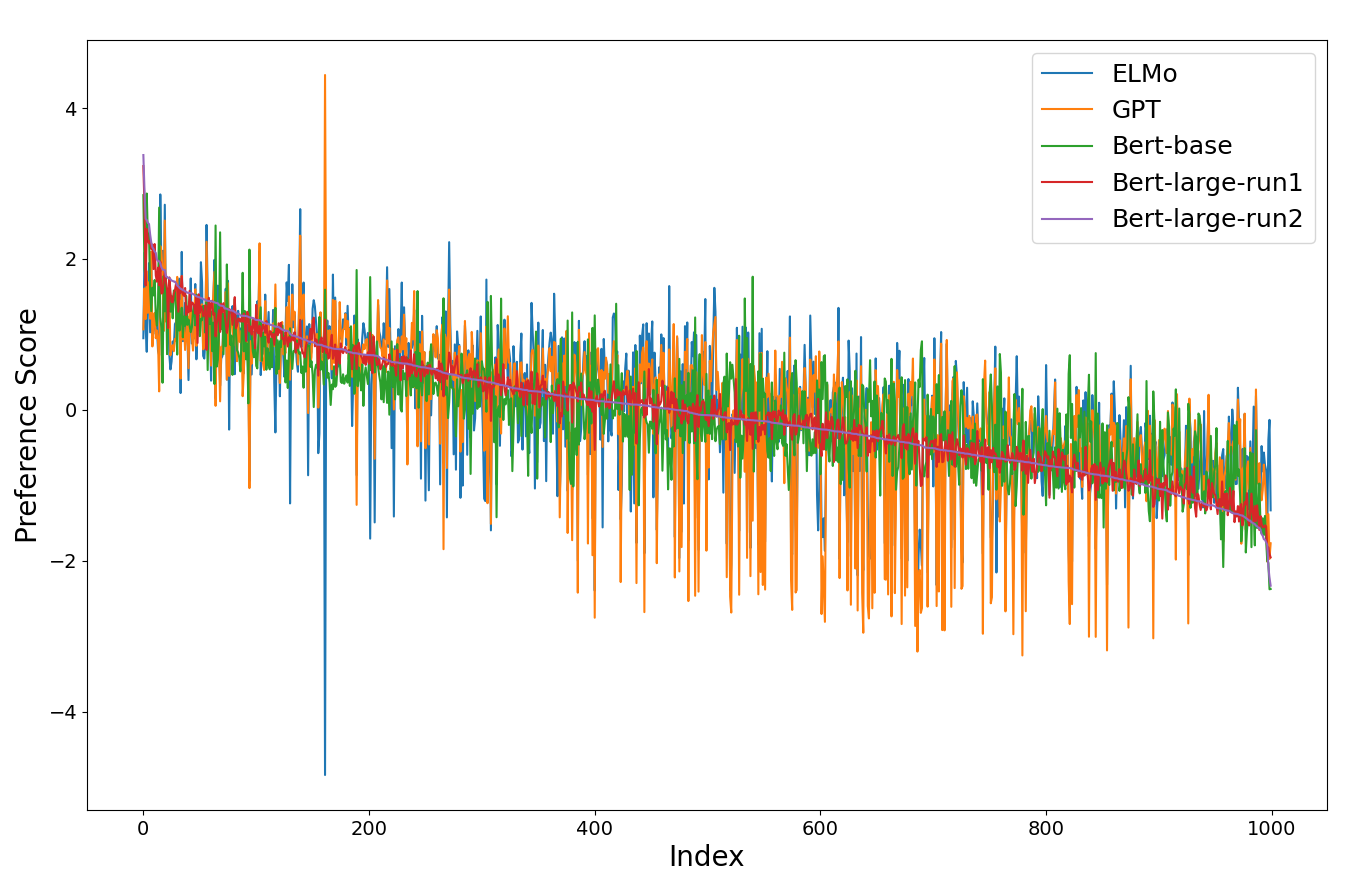}
\caption{Preference score of five cases of two-tower language models. The index of the word is sorted so that the preference score is in descending order concerning Bert-large-run2 case. The score of the most frequent 1000 words is plotted in this figure.}
\label{prefscore}
\end{figure}

\textbf{Same network configuration}: pattern of preference is very stable if we use the same network configuration and vary little according to the random seed for initialization and training. Fig. \ref{prefscore} red and purple lines show the results of two runs of the same Bert-large network configuration with a different random seed. For simplicity, we only show the result for the most frequent thousand tokens. The x-axis represents the token index and the y-axis represents the corresponding preference score. Note that the token index is re-arranged with a descending order of preference score according to the Bert-large-run2 case. We can see that the preference scores of two instances of Bert-large (run1 and run2) follow almost the same trend. We can calculate the correlation between two $s$ vectors for the top 5000 words (to suppress noises introduced by rare words). The correlation can be as high as 0.94. 

\textbf{Different network configuration within same model type}: then, we compare two instances within the same model type but with different network configurations, namely Bert-base and Bert-large. These results are shown in Fig. \ref{prefscore} green and purple lines. It can be seen that the preference score of Bert-base and Bert-large strongly correlate with each other. The correlation between the top 5000 words of two $s$ vectors is 0.72.

\textbf{Different model types}: in this test, we compare two instances among different model types, namely Bert-large, ELMo, and GPT. The comparison result is shown in Fig. \ref{prefscore}. The purple line represents Bert-large; the blue line represents the preference score of ELMo and the orange line represents GPT. It can be seen that the preference score of Bert-large, ELMo, and GPT still correlate to some extent. The correlation between Bert-large and ELMo is 0.57, and between Bert-large and GPT is also 0.57. The correlation is also calculated for the top 5000 words.

\textbf{Pair-wise Correlation}: we conclude the pair-wise correlation among the five cases we've experimented with in Table \ref{correlation}. In light of the stable positive correlation, it can be concluded that there exists a stable pattern of preference scores among two-tower language models.

\begin{table*}[h]
\caption{Pair-wise correlation among the five cases we've experimented on, calculated on the most frequent 5000 words.}
\label{correlation}
\begin{center}
\begin{tabular}{|l|ccccc|}\hline
Model Type      & ELMo &	GPT &	Bert-base &	Bert-large-run1 & Bert-large-run2\\
\hline
ELMo            & 1.0  &  0.91  &  0.27       &       0.57      & 0.61          \\
GPT             & 0.91 &  1.0   &  0.26       &       0.57      & 0.63          \\
Bert-base       & 0.27 &  0.26  &  1.0        &       0.72      & 0.62          \\
Bert-large-run1 & 0.57 &  0.57  &  0.72       &       1.0       & 0.94          \\
Bert-large-run2 & 0.61 &  0.63  &  0.62       &       0.94      & 1.0           \\

\hline
\end{tabular}
\end{center}
\end{table*}

\subsection{Part-of-speech Preference Pattern}

Since the two-tower language model exhibits a stable preference pattern, some tokens are more predictable by one tower than the other. Then, it is natural to ask what kind of tokens are likely to be better predicted by the primary tower and what by the secondary tower. We study this question from a part-of-speech (POS) point of view.

As a first step, we need to assign a POS tag to each token. Even though, in theory, each token may have multiple possible POS tags according to the context. For simplification, we assign the most frequent POS tag to each token in the vocabulary. We pick a subset of our text corpus and use Stanza \citep{qi2020stanza} to generate a POS tag for each token. Then, we go through the vocabulary and assign the most frequent POS tag to each token. As a result, we can now link each preference score calculated for each token to a POS tag. 

\begin{figure}[t]
\centering
\includegraphics[width=1.0\linewidth]{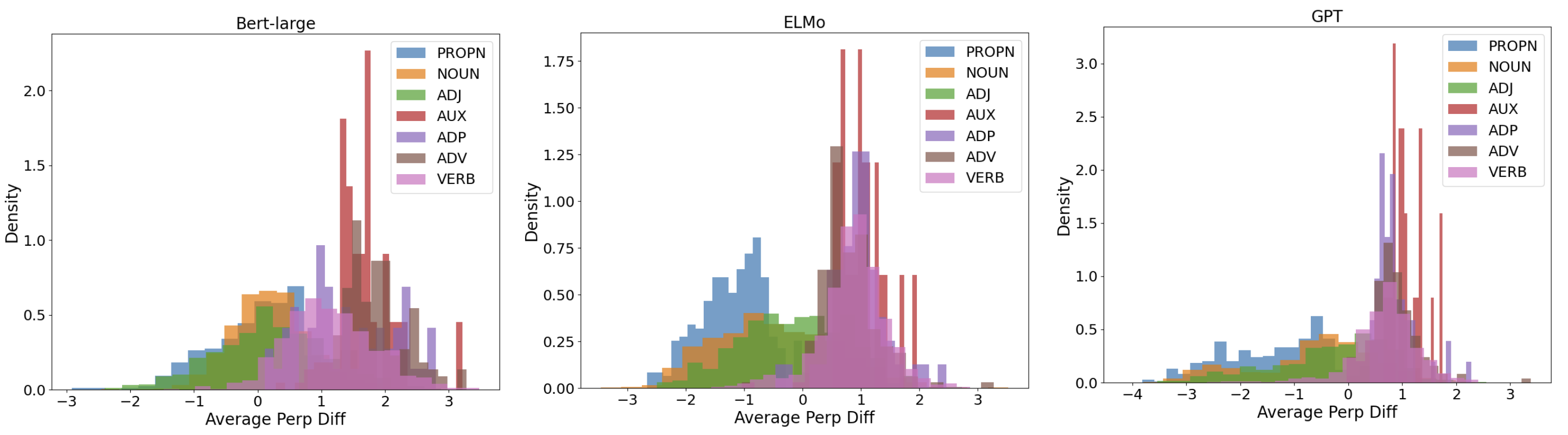}
\caption{Preference score distribution with different POS tags, from left to right, shows Bert-large, GPT, and ELMo cases. The X axis is the preference score, and the Y axis is the normalized distribution.}
\label{POS_dist}
\end{figure}

Fig. \ref{POS_dist} shows the preference score distribution for 7 of the picked POS tags, namely: proper noun (PROPN), noun (NOUN), adjective (ADJ), auxiliary (AUX), adposition (ADP), adverb (ADV), verb (VERB). Note that we use the universal dependency \citep{nivre2016universal} convention for the POS tags. From left to right, three figures represent the Bert-large, GPT, and ELMo cases. It can be seen that VERB, ADP, ADV, and AUX are constantly preferred by the primary tower, PROPN is preferred by the secondary tower, while NOUN and ADJ are more spread out across the two towers. This phenomenon is consistent within all three model types.

\subsection{ICA Component Preference Pattern}

We tested the distribution of preference scores for each POS category in the last section. In this section, we want to study tower preference in a more semantically meaningful manner.

To look in this direction, we follow the idea proposed in \cite{isomura2021achievability}. We first use a principal component analysis (PCA) plus independent component analysis (ICA) method to extract independent language components. Then, we can study the language preference score of each independent language component.

For a subset of word vectors in our corpus $X = [x_1, x_2, ..., x_n]$, we can use PCA to reduce the dimension of word vectors:

\begin{equation}
\begin{split}
& U, S, V = \mathrm{SVD} (X^TX/n) \\
& P = X \cdot V_d \cdot \sqrt{S_d}
\end{split}
\end{equation} 

where $P$ is the corpus with the word vectors reduced to dimension $d$. $\mathrm{SVD}$ is the operator for singular value decomposition. $S_d$ represents the largest $d$ eigenvalues of the covariance matrix $X^TX/n$, n represents the number of tokens, and $V_d$ represents the $d$ eigenvectors corresponding to the largest $d$ eigenvalues. We choose $d$ to be 128 in our study.

After that, we perform ICA analysis on the corpus of word vectors $P$ with reduced dimension $d$. We use Amari's learning rule \citep{amari1995new} to do ICA:

\begin{equation}
\begin{split} \label{ica}
& W_{i+1} = W_m + lr \cdot (I - g(Y_i)Y_i^T/n) \cdot W_i \\
& Y_i = W_i \cdot P
\end{split}
\end{equation}

where $lr$ represents the learning rate (0.01 in our case). $n$ represents the number of tokens. $i$ represents the iteration step. $P$ is the corpus with the word vectors reduced to dimension d. $g(y)$ is the nonlinear activation function:

\begin{equation}
g(y) = \mathrm{sgn}(y) \cdot (1 - \mathrm{exp}(-\sqrt{2} |y|)/2
\end{equation}

where $\mathrm{sgn}$ is the sign function. 

In Eq. \ref{ica}, we iteratively update the source separation matrix $W$ (initialized as a unit matrix $I$) until converge. The resulting $Y$ then becomes the corpus of word vectors with independent language components. 

\begin{figure}[t]
\centering
\includegraphics[width=1.0\linewidth]{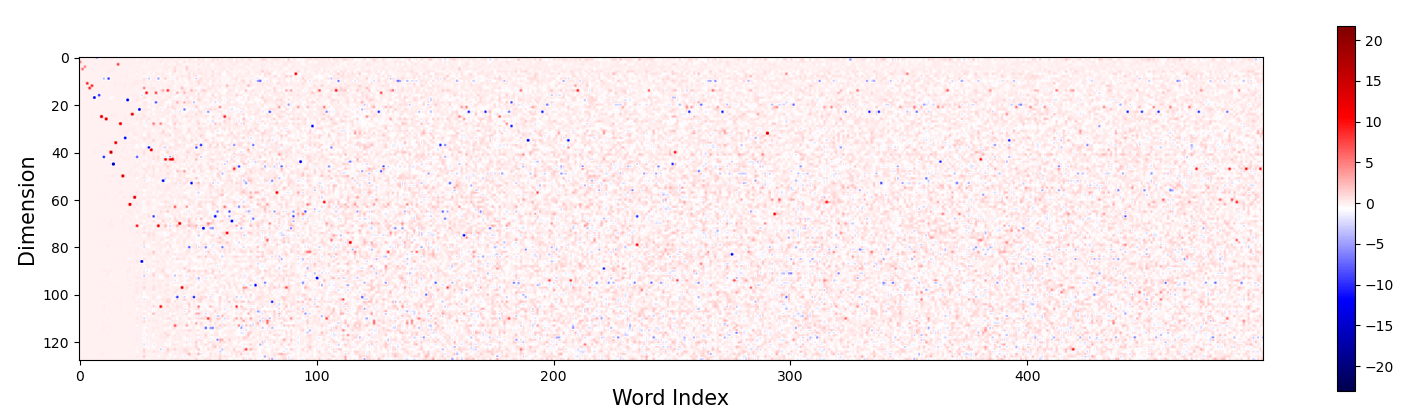}
\caption{Matrix showing the pattern of sparse coding after ICA. The most frequent 500 words are shown in the figure. The X-axis represents the word index, and the Y-axis represents the 128 dimensions of the word vector after ICA. The value of the word vector is represented in color.}
\label{ICA_rep}
\end{figure}

\begin{figure}[t]
\centering
\includegraphics[width=0.9\linewidth]{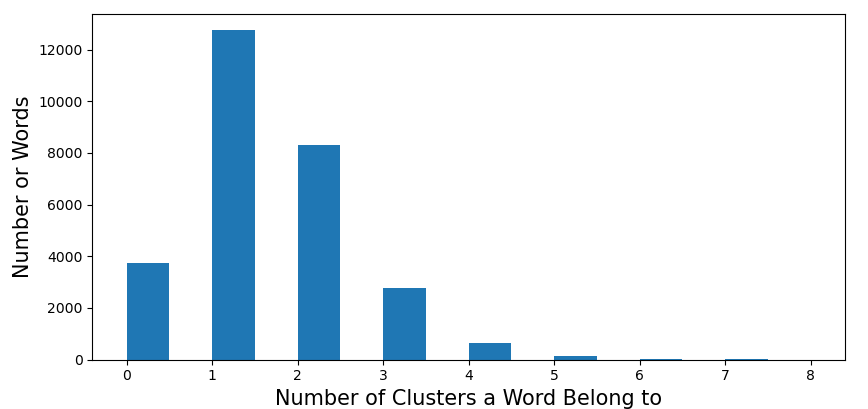}
\caption{Distribution of the number of words that belongs to a certain number of word clusters.}
\label{word_cluster}
\end{figure}

\begin{figure}[t]
\centering
\includegraphics[width=1.0\linewidth]{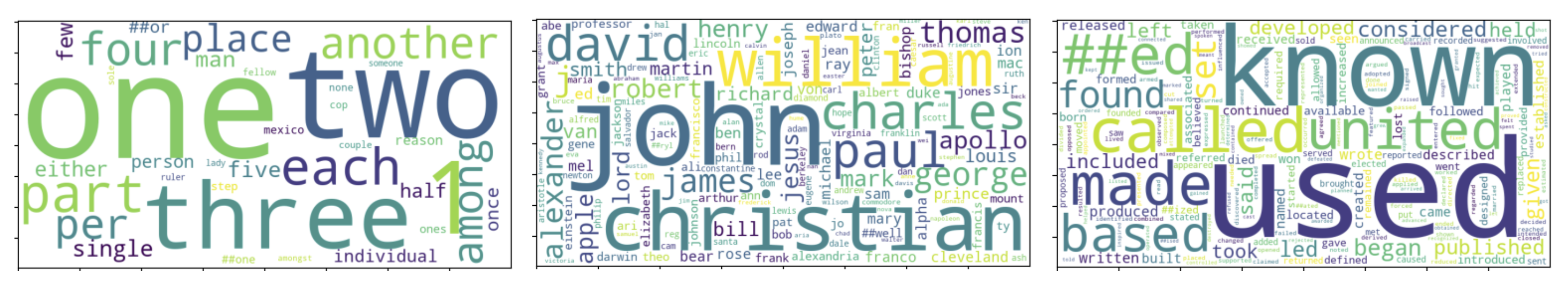}
\caption{Examples of typical word clusters in the form of the word cloud. The size of the word is proportional to its frequency.}
\label{wrdcld}
\end{figure}

We notice that, after the ICA step, the representations of the words become sparse. We plot the representation after ICA in Fig. \ref{ICA_rep}. We can see from the figure that most of the values (around 80\%) are within the range of -1 and 1, with a small fraction of the values being quite high. We can define a word cluster as a group of words whose value at a certain dimension is either larger than 2.5 or smaller than -2.5. With this definition, we can calculate how many clusters each word may correspond to. The result is shown in Fig. \ref{word_cluster}. It can be seen that nearly 45\% of the words correspond to only one cluster. Around 13\% of the words don't belong to any cluster. Each word belongs to an average of 1.45 clusters. We noticed that this method is able to get semantic meaningful word groups. Three examples of the word groups are shown in Fig. \ref{wrdcld} in the form of a word cloud. The font size of the word is proportional to its frequency. As can be seen, word group one is associated with the concept of numbers. Word group two is related to the concept of names. Word group three is about the past tense.

As a next step, we can check the average preference score for each word group. The result is shown in Table \ref{wrdcld_prefscore}. The first column lists selected word groups, which are represented by 5 typical words which belong to this cluster. We calculate the average preference score of all the words in the word group for three of our studied models, namely Bert-large, ELMo, and GPT. We show the result of both the preference score (in the column of the pref-score) and the rank number for reference. As can be read from the table, the self-emerged preference pattern is also stable at the level of the word group. One word group which has a higher preference score for one kind of model also tends to have a higher preference score for another kind of model. 

Another interesting trend that can be seen from the table is that syntactic words which have less semantic meaning tend to have larger preference scores. In contrast, words with more solid semantic meaning tend to have smaller preference scores. This result implies that the separation of syntax and semantics is not only intuition but may also be a self-emerging phenomenon in a two-tower language model.

\begin{table*}[h]
\caption{Preference score and its rank of typical word clusters for Bert-large, ELMo, and GPT, respectively.}
\label{wrdcld_prefscore}
\begin{center}
\begin{tabular}{|l|cc|cc|cc|}\hline
Typical Words  & \multicolumn{2}{|c|}{Bert-large}      & \multicolumn{2}{|c|}{ELMo}      & \multicolumn{2}{|c|}{GPT}      \\
\cline{2-7}
                & pref-score & rank & pref-score & rank & pref-score & rank \\
\hline
which, who, what, where, when                       & 2.16  &  1  &  0.79 & 6 &  0.81   & 5   \\
but, only, than, however, because                   & 2.16  &  2  &  0.98 & 2 &  1.11   & 2   \\
also, not, however, both, then                      & 1.82  &  3  &  0.83 & 4 &  0.86   & 4   \\
is, was, were, be, been                             & 1.77  &  4  &  1.16 & 1 &  1.12   & 1   \\
can, may, would, could, will                        & 1.46  &  5  &  0.94 & 3 &  0.85   & 5   \\
only, especially, simply, directly, probably        & 1.38  &  6  &  0.31 & 11 &  0.31  & 10   \\
one, two, three, each, another                      & 1.28  &  7  &  0.41 & 10 &  0.48  & 7 \\
other, more, later, same, different                 & 1.25  &  8  &  0.67 & 8  &  0.45  & 9   \\
him, them, us, men, whom                            & 1.08  &  9  &  0.78 & 7  &  1.02  & 3   \\
\#\#ing, being, using, having, following            & 0.98  &  10  & 0.56 & 9  &  0.28  & 11   \\
used, called, made, based, known                    & 0.96  &  11  & 0.81 & 5  &  0.60  & 6   \\
may, year, season, march, day                       & 0.70  &  12  & 0.14 & 14 &  0.37  & 9   \\
\#\#s, \#\#ed, \#\#ly, \#\#ing, \#\#ism             & 0.65  &  13  & -1.16& 21  &  -1.78  & 21   \\
new, high, general, early, different                & 0.43  &  14  & 0.26 & 12 &  0.20  & 12   \\
people, human, life, home, children                 & 0.35  &  15  & 0.06 & 15 &  0.04  & 14   \\
john, christian, william, david, charles            & 0.30  &  21  & -0.99& 21 &  -0.85  & 19   \\
film, book, law, history, theory                    & 0.22  &  16  & 0.24 & 13 &  0.05  & 13   \\
cultural, language, music, art, song                & 0.03  &  17  & -0.40 & 17 &  -0.36  & 16   \\
war, people, government, military, president        & 0.02  &  18  & -0.58 & 19 &  -0.62  & 18   \\
head, cell, hand, body, brain                       & -0.16  &  19 & -0.17 & 16 &  -0.24  & 15   \\
metal, air, ice, iron, stone                        & -0.28  &  20 & -0.57 & 18 &  -0.47  & 17   \\
world, states, american, british, french            & -0.69  &  21 & -1.05 & 20 &  -1.37  & 20   \\

\hline
\end{tabular}
\end{center}
\end{table*}

\section{Discussion}

Language is the product of the human brain. A self-emerged pattern of preference in the two-tower language model may be related to various discoveries of the self-emerged pattern of preference in the human brain. One famous example is the cerebral dominance hypothesis \citep{brown1975hypothesis}. It suggests that one cerebral hemisphere is primarily responsible for certain types of functions, while the other functions would be controlled by the other cerebral hemisphere. A second example is the two cortical pathways hypothesis in human vision \citep{mishkin1983object}. This hypothesis suggests there exist two separate pathways: the ventral pathway, which is in charge of the visual identification of objects, and the dorsal pathway, which is in charge of the localization of objects. Another hypothesis exists for speech perception. For example, Hickok et al. \citep{hickok2004dorsal} hypothesized that there exists a ventral stream for mapping sound onto meaning, and a dorsal stream for mapping sound onto articulatory-based representations.

\section{Conclusion}

In this paper, we present our study of a two-tower language model setting, where two symmetrical language models with the same network configuration are trained side-by-side to do language modeling tasks. We find that, even though the language models start symmetrical, preference for certain language functions emerges after training. We define the preference score of tokens and find a significant correlation between the scores across model instances with different network settings. We also discover that verbs, adverbs, adpositions, and auxiliary tokens are usually better modeled by the primary language model tower, proper nouns by the secondary language model tower. While adjectives and nouns are more spread out. This preference is rather stable across different network configurations and types, indicating it is an intrinsic property of natural language. We also discovered the self-emerged separation of syntax and semantics in a two-tower language model. We suggest that more work should be invested into understanding the nature of our language so that we can engineer smarter and better language processing techniques in the future.

\section{Supplementary}

\subsection{Preference Score and Word Frequency}

This research discovers a novel phenomenon in which preferences emerge spontaneously over diverse words when a two-tower language model is used. We define preference scores and demonstrate them as qualitatively stable across different types of deep neural networks. We also find the preference score correlates with POS types and ICA word groups, and find out that the primary tower prefers to model VERB, ADP, ADV, AUX, and more syntactic word groups, while the secondary tower prefers to model PROPN and more semantic word groups. 

Another salient feature of the word that is known to be correlated with word function is word frequency. As we know, functional words tend to have a higher frequency, and more frequent words tend to have more solid semantic meaning. Then we ask, are we just rediscovering word frequency with a two-tower language model? The answer is no.

\begin{figure}[t]
\centering
\includegraphics[width=1.0\linewidth]{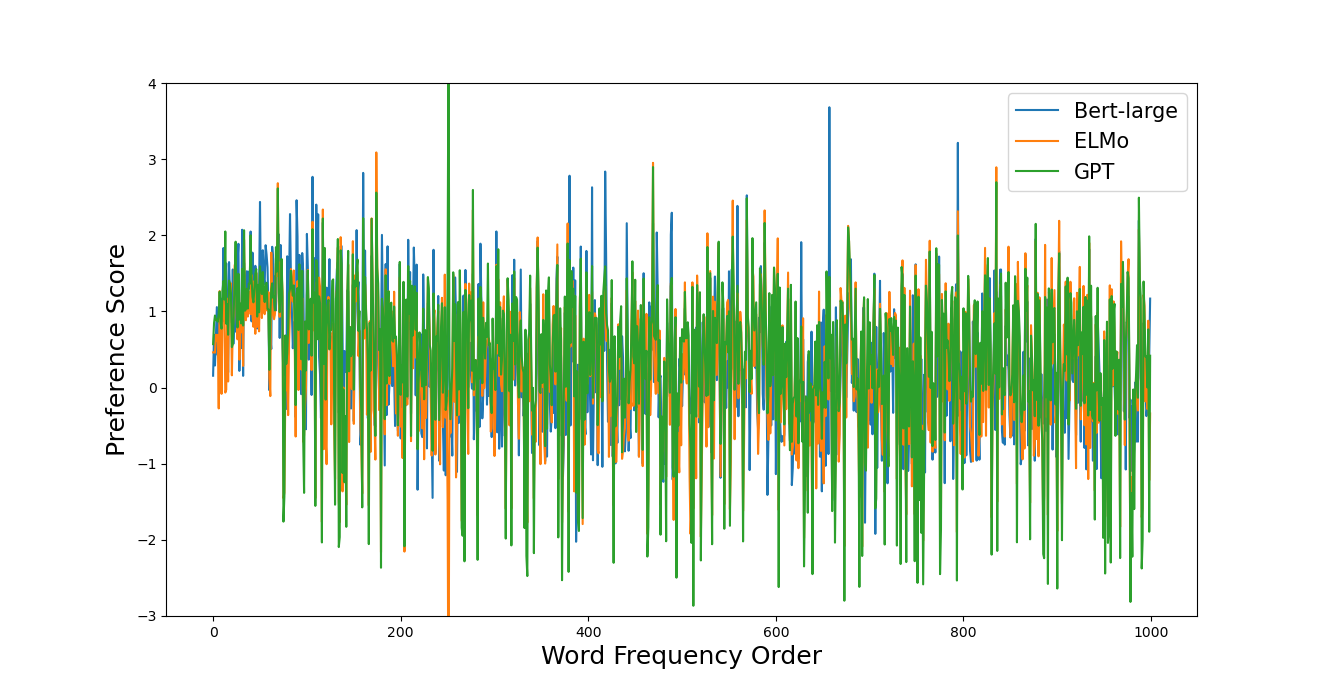}
\caption{Word preference score v.s. word index which is sorted in descending order. Blue, orange, and green lines are for Bert-large, ELMo, and GPT respectively.}
\label{wrdord_freq}
\end{figure}

Fig. \ref{wrdord_freq} shows the preference score v.s. word index sorted with descending word frequency of three types of the tow-tower language model, namely Bert-large (blue), ELMo (orange), and GPT (green). The plot shows the most frequent 1000 words for simplicity. It can be seen from the figure that, except for the first several tens of words, preference score and word frequency have no obvious correlation.

\begin{table*}[h]
\caption{Pair-wise Spearman's rank correlation among Bert-large, ELMo, GPT, and frequency rank, calculated on the most frequent 5000 words.}
\label{SpearmanCorr}
\begin{center}
\begin{tabular}{|l|cccc|}\hline
Model Type      & ELMo &	GPT & Bert-large  & Frequency Rank\\
\hline
ELMo            & 1.0  &  0.93  &  0.56       &       0.16      \\
GPT             & 0.93 &  1.0   &  0.65       &       0.22      \\
Bert-large      & 0.56 &  0.65  &  1.0        &       0.18      \\
Frequency Rank  & 0.16 &  0.22  &  0.18       &       1.0       \\

\hline
\end{tabular}
\end{center}
\end{table*}

We also check the correlation quantitatively. Table \ref{SpearmanCorr} shows Spearman's rank correlation among Bert-large, ELMo, GPT preference score rank, and word frequency rank. It can be seen that the correlation between word frequency rank and preference score rank of language models is significantly smaller than the correlation of preference score rank between the two language models. These results confirm that the preference score pattern described in this paper cannot be solely explained by word frequency pattern.

\bibliographystyle{abbrvnat}
\bibliography{TTLM_main}

\end{document}